\documentclass[letterpaper]{article} 
\usepackage{aaai2027}  
\usepackage[hyphens]{url}  
\usepackage{graphicx} 
\usepackage{natbib}  
\usepackage{caption} 
\usepackage{amsmath}
\usepackage{amssymb}

\usepackage[hyphens]{url}  
\usepackage{graphicx} 
\usepackage{natbib}  
\usepackage{caption} 
\usepackage{algorithm}
\usepackage{algorithmic}
\usepackage{multirow}
\usepackage{colortbl}
 
\usepackage{subcaption} 
\usepackage{newfloat}
\usepackage{listings}
\DeclareCaptionStyle{ruled}{labelfont=normalfont,labelsep=colon,strut=off} 
\floatstyle{ruled}
\newfloat{listing}{tb}{lst}{}
\floatname{listing}{Listing}

\usepackage{booktabs}

\title{
From Spatial Semantics to Temporal Context: Leveraging Gaze Trajectory for Weakly Supervised Medical Image Segmentation
}
\author{
    Shaoxuan Wu\textsuperscript{\rm 1},
    Xiao Zhang\textsuperscript{\rm 1}\corresponding,
    Xiaodi Zhao\textsuperscript{\rm 1}
    Yunzhi Tian\textsuperscript{\rm 1}
    Yilin Tang\textsuperscript{\rm 1}
    Jun Feng\textsuperscript{\rm 1}\corresponding
}
\affiliations{
    \textsuperscript{\rm 1}College of Computer Science, Northwest University, Xi’an, China\\
    xiaozhang@nwu.edu.cn,
    fengjun@nwu.edu.cn
}

\begin{document}

\maketitle

\begin{abstract}
Medical image segmentation heavily depends on labor-intensive and time-consuming pixel-level annotations. Eye tracking offers a cost-effective solution that can be naturally integrated into clinical workflows. 
Recorded by eye trackers, gaze conveys the spatial regions of clinicians' attention through fixations and the temporal context of clinicians' progressive visual perception from trajectories.
Nevertheless, effective modeling of temporal trajectories remains challenging, and noise in gaze caused by exploratory fixations greatly limits segmentation performance.
To overcome these limitations, we propose the Trajectory-guided Uncertainty-aware Network (TrailNet), which exploits gaze-supervised medical image segmentation from spatial semantics modeling to temporal context by jointly leveraging fixations and trajectories.
Specifically, the proposed trajectory-guided spatio-temporal encoder models temporal context and establishes complementary interactions with image spatial semantics to strengthen target perception. Furthermore, the multi-scale uncertainty decoder leverages category mutual-exclusivity constraints to produce deterministic predictions and mitigate supervision uncertainty induced by noise. 
To enable gaze-free inference, we further introduce a cycle distillation strategy that transfers feature-level knowledge via teacher-student networks.
Experimental results on two public datasets demonstrate that TrailNet outperforms state-of-the-art methods, achieving Dice scores of 81.25\% and 81.85\%, respectively.
\end{abstract}

\begin{links}
\end{links}

\section{Introduction}
Medical image segmentation focuses on localizing regions of interest and constitutes a fundamental technique for early disease screening and precision treatment \cite{zhang2026decoding}. However, the strong performance of deep learning-based segmentation models typically relies on large-scale and high-quality pixel-level annotations. Producing such annotations requires substantial domain expertise and clinical experience, making the process labor-intensive and time-consuming, thereby hindering the deployment of models in real-world clinical scenarios \cite{Shen2023survey,zhang2025generative}. To alleviate annotation costs, weakly supervised medical image segmentation has been widely explored by leveraging bounding boxes \cite{11049003}, points \cite{10918702}, or scribbles \cite{XU2026131642}. Nevertheless, such paradigms not only require clinicians to perform additional annotation efforts beyond routine image interpretation, increasing clinical workload, but also encode only static annotation results and fail to capture the diagnostic process of clinicians \cite{GANet}.

\begin{figure}[t]
\centering
\includegraphics[width=0.99\columnwidth]{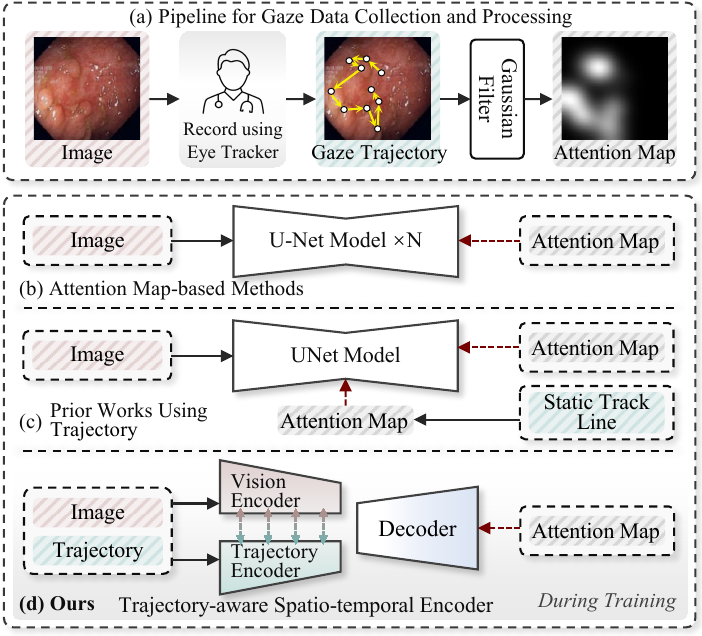}
\caption{(a) Pipeline for gaze data collection and processing. (b) Attention map-based methods discard valuable trajectory information. (c) Previous methods consider trajectories but collapse them into attention maps and neglect the temporal information contained in gaze. (d) Our TrailNet explicitly models trajectories and establishes complementary interactions with image semantics to enhance target perception.}
\label{fig_intro}
\end{figure}

Unlike traditional weak supervision paradigms, gaze records clinicians' visual behaviors during image reading \cite{Dermatologist,yu2026learning,252590}. Moreover, gaze can be automatically collected using eye-trackers and seamlessly integrated into routine clinical workflows with minimal additional costs \cite{RadioTransformer,11079704}. It contains two valuable types of information. Gaze locations reflect the spatial regions of clinicians' attention \cite{11079704}, while gaze trajectories record the progressive process of observing, analyzing, and confirming targets, providing temporal context \cite{GradTrack}.
However, existing gaze-supervised methods mainly focus on the spatial localization capability of gaze \cite{EGViT}. As shown in Fig.~\ref{fig_intro}(a), most methods first transform discrete gaze coordinates into attention maps through Gaussian filtering, which are then utilized for region-level supervision \cite{Xie2024integrating,Wang2023eye}. In gaze-supervised medical image segmentation, as illustrated in Fig.~\ref{fig_intro}(b), mainstream methods typically generate attention maps and train multiple expert models with different thresholds to improve segmentation performance \cite{zhong2024weakly,chen2025gaze,GNAN}. 
Furthermore, as shown in Fig.~\ref{fig_intro}(c), although GradTrack considers fixation order, it still converts trajectories into static representations and attention maps, failing to explicitly model temporal information within gaze trajectories \cite{GradTrack}.

Gaze trajectories encode the temporal context of clinicians’ progressive observation, enabling models to comprehend complex anatomical structures and ambiguous lesion boundaries under sparse weak supervision \cite{GiTNet,GE2025107865}. In particular, gaze trajectories capture sequential diagnostic behaviors, whereas medical images provide pathological spatial semantics \cite{NEURIPS2025_be7bc8aa,10943557,pathologists}. The two modalities deliver complementary cues for target representation learning. 
\textit{Therefore, gaze extends beyond spatial supervision, with trajectories providing valuable temporal context that can be explicitly modeled and effectively integrated with image semantics to further exploit the supervisory potential.}
Nevertheless, effective utilization of gaze trajectories still faces challenges. First, the temporal context embedded in trajectories exhibits complex sequential structures, making it difficult to model effectively \cite{11444425}. Second, image spatial semantics and gaze temporal context are heterogeneous, and their complementary cross-modal interaction remains under-explored \cite{WANG2025111350,10839445}. Finally, gaze inevitably contains noisy exploratory fixations that deviate from pathological regions and lack precise boundaries, leading to inherent supervision uncertainty.

To address these challenges, we propose Trajectory-guided Uncertainty-aware Network (TrailNet) for gaze-supervised medical image segmentation. As depicted in Fig.~\ref{fig_intro}(d), TrailNet explicitly models gaze temporal context and establishes complementary interactions with image spatial semantics.
Specifically, we first introduce the Trajectory-guided Spatio-temporal Encoder (TSE). Within the TSE, the trajectory perception block captures temporal context from the trajectory, while a context-semantic interaction block explores the complementary relationship between spatial semantics and temporal context, thereby enhancing target perception during the encoding.
Furthermore, we propose the Multi-scale Uncertainty Decoder (MUD), which leverages category mutual-exclusivity constraints to encourage stable predictions and alleviate supervision uncertainty caused by noise. Finally, considering gaze as a training-time privileged modality, we introduce the Cycle Distillation Strategy (CDS) to transfer trajectory feature knowledge to the student model, enabling reliable gaze-free inference while retaining precise structural perception of lesions.

In summary, our contributions are as follows:
\begin{itemize}
\item We propose TrailNet for gaze-supervised medical image segmentation. The devised TSE captures the temporal context of clinicians’ progressive perception from gaze trajectories and enables complementary interactions with image spatial semantics to boost the network’s perception of target structures.
\item Within TrailNet, we develop the MUD for uncertainty-aware decoding, where multi-scale category mutual-exclusivity constraints drive deterministic predictions to mitigate gaze noise and uncertain supervision and improve segmentation reliability.
\item We develop the CDS, which implements feature-level distillation via teacher-student networks and cycle distillation constraints to enable gaze-free inference.
\item Experiments on two public datasets demonstrate that TrailNet outperforms existing weakly supervised medical image segmentation methods.
\end{itemize}

\begin{figure*}[t]
\centering
\includegraphics[width=0.99\textwidth]{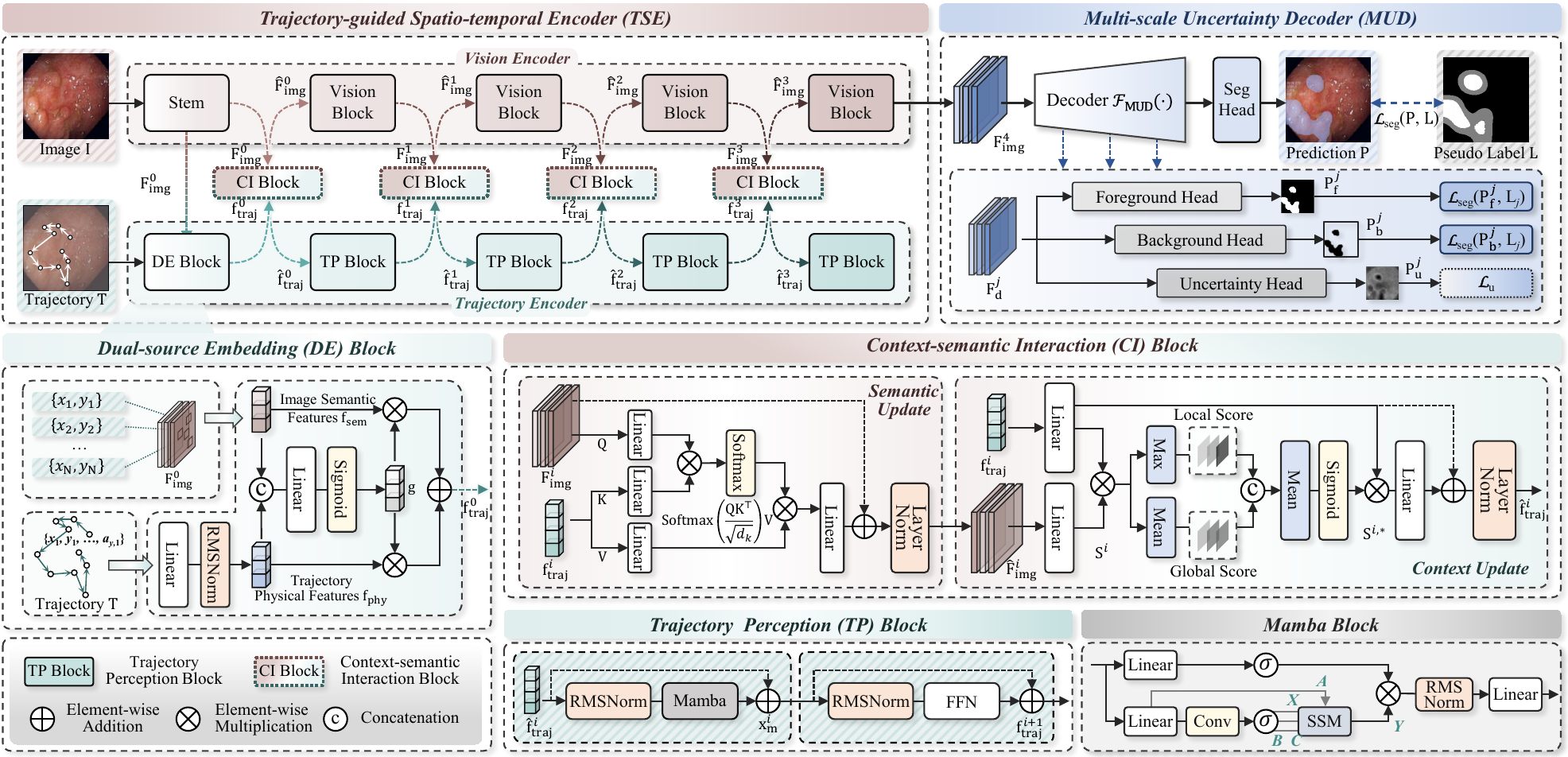}
\caption{Overview of TrailNet, which consists of three core modules. TSE achieves Context-semantic Interaction of spatial image semantics and temporal trajectory features. MUD alleviates gaze noise via category mutual-exclusivity constraints. 
In TSE, DE produces initial trajectory embeddings, TP models trajectory temporal context through a two-layer residual Mamba structure, and CI implements progressive mutual enhancement between spatial semantics and trajectory context.
}
\label{fig_method1}
\end{figure*}

\section{Methodology}

\subsection{Method Overview and Preliminaries}

The proposed TrailNet consists of three core modules: the Trajectory-guided Spatio-temporal Encoder (TSE), Multi-scale Uncertainty Decoder (MUD), and Cycle Distillation Strategy (CDS). The overall architecture is shown in Fig.~\ref{fig_method1}. Our core objective is to model gaze temporal context and achieve gaze-supervised medical image segmentation via complementary interactions with image spatial semantics. 
Given an input image $\mathbf{I} \in \mathbb{R}^{H \times W \times C}$, where $H$, $W$, and $C$ correspond to the image height, width, and channel count. The corresponding gaze trajectory is a temporally ordered fixation sequence $\mathbf{T} = \{(x_t, y_t,d_t,v_{x,t},v_{y,t}, a_{x,t}, a_{y,t})\}_{t=1}^{N}$. $(x_t, y_t)$ stands for the coordinate of the $t$-th fixation, $d$ denotes the fixation duration, $v$ and $a$ denote the velocity and acceleration, and $N$ is the total number of fixations.

First, the image and gaze trajectory are jointly fed into the TSE for extraction and fusion of image spatial semantics and temporal context features of the trajectory. The fused features are then forwarded to the MUD to produce the segmentation probability map $\mathbf{P} \in [0,1]^{H \times W}$. During training, pseudo-labels generated from gaze signals are adopted as weak supervision. Benefiting from the CDS, only images are required for segmentation during inference.
Following \citet{GNAN}, for the fixation set $\textbf{F}$, a 2D Gaussian filter is applied to generate the attention map $\mathbf{A}$, as illustrated in Fig.~\ref{fig_method2}. We set the foreground threshold $\tau_{\text{f}}$ and background threshold $\tau_{\text{b}}$ to classify pixels into foreground, background, and uncertain regions, yielding pseudo-labels $\mathbf{L} \in \{1, 0, -1\}^{H \times W}$, where $1$, $0$ and $-1$ denote foreground, background and uncertain regions, respectively. Uncertain pixels receive no direct supervision during training.

\subsection{Trajectory-guided Spatio-temporal Encoder (TSE)}
The TSE explicitly models temporal context and establishes complementary interactions with image spatial semantics to strengthen target perception. Given input image $\mathbf{I}$, the vision encoder first downsamples via a stem layer to produce $\mathbf{F}_{\text{img}}^0$. Successive vision blocks extract multi-scale image features $\{\mathbf{F}_{\text{img}}^i\}_{i=1}^{4}$, where $\mathbf{F}_{\text{img}}^i \in \mathbb{R}^{H_i \times W_i \times D_i}$ represents the $i$-th level image feature map. Meanwhile, trajectory $\mathbf{T}$ enters the trajectory encoder: the Dual-source Embedding (DE) Block initializes trajectory embeddings, and stacked Trajectory Perception (TP) Blocks yield multi-scale temporal trajectory features $\{\mathbf{f}_{\text{traj}}^i\}_{i=1}^{4}$ with $\mathbf{f}_{\text{traj}}^i \in \mathbb{R}^{N \times D_i}$. Multi-level image and trajectory features are then fed into the Context-semantic Interaction (CI) Block for feature fusion.

\subsubsection{Dual-source Embedding (DE) Block.}
The DE Block utilizes image semantic and trajectory physical features to build initial trajectory embeddings.
Given trajectory $\mathbf{T} \in \mathbb{R}^{N \times 7}$, whose seven dimensions encode coordinates, duration, velocity, and acceleration, a linear layer with RMSNorm extracts trajectory physical features $\mathbf{f}_{\text{phy}} \in \mathbb{R}^{N \times D}$. Meanwhile, bilinear interpolation samples semantic features $\mathbf{f}_{\text{sem}} \in \mathbb{R}^{N \times D}$ from $\mathbf{F}_{\text{img}}^0$ at fixation locations.
We adopt a concise fusion strategy to obtain the initial trajectory embedding $\mathbf{f}_{\text{traj}}^0$:
\begin{equation}
\begin{aligned}
\mathbf{g} &= \sigma\big(\mathrm{Linear}\big(\mathrm{Concat}(\mathbf{f}_{\text{phy}}, \mathbf{f}_{\text{sem}})\big)\big), 
\end{aligned}
\end{equation}
\begin{equation}
\begin{aligned}
\mathbf{f}_{\text{traj}}^0 &= \mathbf{g} \odot \mathbf{f}_{\text{phy}} + (1 - \mathbf{g}) \odot \mathbf{f}_{\text{sem}},
\end{aligned}
\end{equation}
where $\sigma(\cdot)$ is the Sigmoid activation, $\odot$ denotes element-wise multiplication, and $\mathbf{g}$ is the adaptive feature-wise gating coefficient.

\subsubsection{Context-semantic Interaction (CI) Block.}
The CI Block achieves progressive mutual enhancement of bimodal features: temporal context strengthens spatial semantics, while semantics suppress trajectory noise.
For input image spatial semantics $\mathbf{F}_{\text{img}}^i$ and trajectory temporal context $\mathbf{f}_{\text{traj}}^i$, linear layers build a cross-attention Query, Key, and Value, where $\mathbf{F}_{\text{img}}^i$ acts as Query and $\mathbf{f}_{\text{traj}}^i$ provides Key and Value. The cross-attention is computed as:
\begin{equation}
\hat{\mathbf{F}}_{\text{img}}^i = \text{LN}(\mathrm{Linear}(\mathrm{Softmax}\left( \frac{\mathbf{Q} \mathbf{K}^\top}{\sqrt{d_k}} \right) \mathbf{V}) + \mathbf{F}_{\text{img}}^i),
\end{equation}
with $d_k$ the feature dimension. The result is reshaped back to spatial form, highlighting regions guided by trajectory priors.
Next, we generate adaptive weights to denoise trajectory features. Linear projections transform the two modalities into $\hat{\mathbf{F}}_{\text{img}}^{i,*}$ and $\mathbf{f}_{\text{traj}}^{i,*}$, whose element-wise product forms matching score matrix $\mathbf{S}^i$. Calibration weights $\mathbf{S}^{i,*}$ are derived via:
\begin{equation}
\mathbf{S}^{i,*} = \sigma\big( \mathrm{Mean}\big( \mathrm{Concat}\big( \max_{d}(\mathbf{S}^i), \min_{d}(\mathbf{S}^i) \big) \big) \big).
\end{equation}

We weight the trajectory features by $\mathbf{S}^{i,*}$ and apply LayerNorm plus linear projection for the final output:
\begin{equation}
\hat{\mathbf{f}}_{\text{traj}}^{i} = \mathrm{LN}\big( \mathrm{Linear}\big( \mathbf{S}^{i,*} \odot {\mathbf{f}}_{\text{traj}}^{i,*} \big)\big).
\end{equation}

\begin{figure}[t]
\centering
\includegraphics[width=0.99\columnwidth]{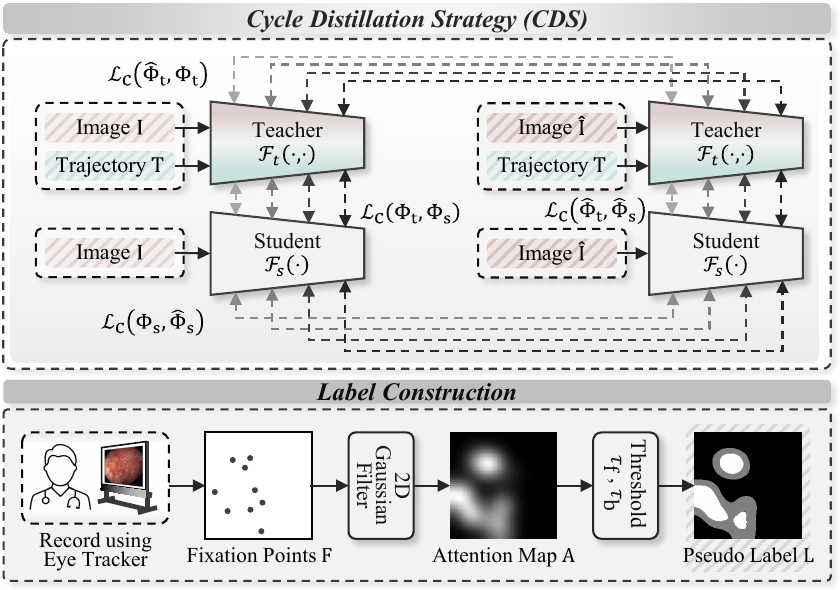}
\caption{CDS constructs cycle distillation constraints with teacher-student networks for feature-level knowledge distillation. The attention map is generated by a 2D Gaussian filter on the fixation, and pseudo-labels are obtained with the thresholds $\tau_{\text{f}}$ and $\tau_{\text{b}}$.
}
\label{fig_method2}
\end{figure}

\subsubsection{Trajectory Perception (TP) Block.}
The TP Block captures the temporal context of the trajectory with a two-layer residual architecture centered on the Mamba state space model.
Given the $i$-th layer trajectory feature $\hat{\mathbf{f}}_{\text{traj}}^{i}$, the TP Block computation follows:
\begin{equation}
\begin{aligned}
\mathbf{x}_{\text{m}}^i &= \mathrm{Mamba}\big(\mathrm{RMSNorm}(\hat{\mathbf{f}}_{\text{traj}}^{i})\big) + \hat{\mathbf{f}}_{\text{traj}}^{i}, 
\end{aligned}
\end{equation}
\begin{equation}
\begin{aligned}
\mathbf{f}_{\text{traj}}^{i+1} &= \mathrm{FFN}\big(\mathrm{RMSNorm}(\mathbf{x}_{\text{m}}^i)\big) + \mathbf{x}_{\text{m}}^i,
\end{aligned}
\end{equation}
where RMSNorm is root mean square normalization, and FFN represents the feed-forward network.
Mamba builds on continuous state space models (SSM) and achieves efficient temporal modeling via input-dependent discretization and selective scan \cite{mamba}. Following Mamba, all state parameters are dynamically generated from input tokens.
For input sequence $\mathbf{z} = \{\mathbf{z}_t\}_{t=1}^N$, the per-step SSM parameters are projected from each input token:
\begin{equation}
\Delta_t, \mathbf{B}_t, \mathbf{C}_t = f_\theta(\mathbf{z}_t),
\end{equation}
where $f_\theta(\cdot)$ denotes a learnable linear projection, $\Delta_t$ is the input-adaptive discretization step size, and $\mathbf{B}_t, \mathbf{C}_t$ are input-conditioned state matrices. The discretized state recurrence is formulated as:
\begin{equation}
\begin{aligned}
\mathbf{h}_t &= \bar{\mathbf{A}}_t \mathbf{h}_{t-1} + \bar{\mathbf{B}}_t \mathbf{z}_t, 
\end{aligned}
\end{equation}
\begin{equation}
\begin{aligned}
\mathbf{y}_t &= \mathbf{C}_t \mathbf{h}_t,
\end{aligned}
\end{equation}
where $\mathbf{h}_t$ denotes the hidden state at time step $t$, and $\bar{\mathbf{A}}_t, \bar{\mathbf{B}}_t$ are discretized transition matrices derived from continuous SSM parameters and $\Delta_t$.

\subsection{Multi-scale Uncertainty Decoder (MUD)}
The MUD alleviates gaze noise and supervision uncertainty via multi-scale uncertainty modeling and category mutual-exclusivity constraints, improving segmentation reliability.
Taking TSE output features as input, the MUD decoder $\mathcal{F}_{\text{MUD}}(\cdot)$ progressively upsamples to generate three-scale decoded features $\{\mathbf{F}_{\text{d}}^j\}_{j=1}^{3}$. At each scale, three $1{\times}1$ convolution heads predict foreground $\mathbf{P}_{\text{f}}^j$, background $\mathbf{P}_{\text{b}}^j$ and uncertainty $\mathbf{P}_{\text{u}}^j$ probability maps respectively. The decoder finally outputs the full-resolution segmentation map $\mathbf{P}$.

Category mutual-exclusivity regularization is enforced on the three probability maps per scale, with the loss defined as:
\begin{equation}
\mathcal{L}_\text{u} = \sum_{j=1}^{3} \left\| \big(\mathbf{P}_{\text{f}}^j \odot \mathbf{P}_{\text{b}}^j + \mathbf{P}_{\text{b}}^j \odot \mathbf{P}_{\text{u}}^j + \mathbf{P}_{\text{u}}^j \odot \mathbf{P}_{\text{f}}^j \big)\odot (1 + \mathbf{A})\right\|_1,
\end{equation}
where $\|\cdot\|_1$ denotes the $\ell_1$ norm and $\mathbf{A}$ is the gaze attention map used to weight foreground regions.
Foreground and background maps at each scale are supervised by gaze-derived pseudo-labels, with loss computed only on certain regions:
\begin{equation}
\mathcal{L}_\text{m} = \sum_{j=1}^{3} \sum_{c \in \{\text{f},\text{b}\}} \mathcal{L}_{\text{seg}}(\mathbf{P}_{c}^j, \mathbf{L}_j),
\end{equation}
where $\mathbf{L}_j$ is the pseudo-label downsampled to the $j$-th scale. The masked cross-entropy $\mathcal{L}_{\text{seg}}(\cdot)$ is formulated as:
\begin{equation}
\begin{aligned}
\mathcal{L}_{\text{seg}}(\mathbf{P}, \mathbf{L})
&= -\frac{1}{\sum_{x,y} \mathbf{M}(x,y)}
\sum_{x,y} \mathbf{M}(x,y) \log \mathbf{P}_{L(x,y)}(x,y),
\end{aligned}
\end{equation}
where $\mathbf{M}(x,y) = \mathbb{I}[L(x,y) \neq -1]$ is a valid pixel mask, and $\mathbb{I}[\cdot]$ denotes the indicator function.
The total MUD loss is thus:
\begin{equation}
\mathcal{L}_{\text{MUD}} = \mathcal{L}_\text{u} + \mathcal{L}_\text{m}.
\end{equation}

\begin{table*}[!t]
    \small
    \centering
    
    \begin{tabular}{
    m{5.7cm} | 
    m{2.4cm}<{\centering}|
    m{1.6cm}<{\centering}|  
    m{2.1cm}<{\centering}| 
    m{2.1cm}<{\centering}|
    m{1.3cm}<{\centering}  }
    
    \toprule
    
    \multirow{2}{*}{\textbf{Method}}
    & \multirow{2}{*}{\textbf{Venue \& Year}}
    & \multirow{2}{*}{\textbf{Supervision}}
    & \textbf{NCI-ISBI}
    & \multicolumn{2}{c}{\textbf{Kvasir-SEG}}
    \\
    
    \cmidrule(lr) {4-4}
    \cmidrule(lr) {5-6}
    & & & \textbf{Dice(\%)$\uparrow$} & \textbf{Dice(\%)$\uparrow$} & \textbf{AT$\downarrow$} \\
    
    \midrule
    U-Net (Ronneberger et al. 2015)
    & MICCAI'15 & Full   & 80.58~$\pm$~0.48  
    & 82.12~$\pm$~1.11  & 18.7 hrs   \\
    nnU-Net \cite{nnUNet}
    & Nat. Methods'21 & Full   & 81.54~$\pm$~0.45  
    & 85.37~$\pm$~0.48  & 18.7 hrs   \\
    TransUNet \cite{TransUNet}
    & MedIA'24 & Full   & 81.02~$\pm$~0.48  
    & 85.05~$\pm$~0.54  & 18.7 hrs   \\
    \midrule
    \rowcolor{gray!20}
    \textbf{TrailNet$_\text{F}$ (Ours)}
    & \textbf{Ours} & Full   & \textbf{82.78}~$\pm$~0.42  
    & \textbf{86.68}~$\pm$~0.52  & 18.7 hrs   \\
    
    \midrule
    BoxInst \cite{BoxInst}
    & CVPR'21 & Box   & 73.78~$\pm$~1.15  
    & 65.72~$\pm$~2.97  & 3.1 hrs   \\
    BoxTeacher \cite{Boxteacher}
    & CVPR'23 & Box   & 75.60~$\pm$~1.15  
    & 73.33~$\pm$~1.30  & 3.1 hrs   \\
    \midrule
    
    PointSup \cite{Pointsup}
    & CVPR'22 & Point   & 73.46~$\pm$~4.71  
    & 73.05~$\pm$~1.64  & 4.8 hrs   \\
    AGMM \cite{AGMM}
    & CVPR'23 & Point   & 73.86~$\pm$~1.26  
    & 75.57~$\pm$~0.84  & 4.8 hrs   \\
    \midrule
    
    CycleMix \cite{CycleMix}
    & CVPR'22 & Scribble   & 73.41~$\pm$~1.09  
    & 76.43~$\pm$~0.65  & 2.6 hrs   \\
    AGMM \cite{AGMM}
    & CVPR'23 & Scribble   & 72.70~$\pm$~1.03  
    & 67.23~$\pm$~1.02  & 2.6 hrs   \\
    ShapePU \cite{ShapePU}
    & MICCAI'22 & Scribble   & 73.06~$\pm$~1.18  
    & 77.26~$\pm$~0.73  & 2.6 hrs   \\
    ScribFormer \cite{ScribFormer}
    & TMI'24 & Scribble   & 74.31~$\pm$~1.29  
    & 75.69~$\pm$~0.48  & 2.6 hrs   \\
    
    \midrule
    U-Net (Ronneberger et al. 2015)
    & MICCAI'15 & Gaze   & 74.75~$\pm$~1.58  
    & 73.74~$\pm$~0.94  & 2.2 hrs   \\
    nnU-Net \cite{nnUNet}
    & Nat. Methods'21 & Gaze   & 77.20~$\pm$~1.03  
    & 74.42~$\pm$~0.92  & 2.2 hrs   \\
    TransUNet \cite{TransUNet}
    & MedIA'24 & Gaze   & 75.46~$\pm$~1.20  
    & 70.38~$\pm$~0.86  & 2.2 hrs   \\
    GazeMedSeg \cite{zhong2024weakly}
    & MICCAI'24 & Gaze   & 77.64~$\pm$~0.57 
    & 77.80~$\pm$~1.02  & 2.2 hrs   \\
    Chen et al. \cite{chen2025gaze} 
    & TMI'25 & Gaze   & 80.53~$\pm$~0.49 
    & 80.78~$\pm$~0.11  & 2.2 hrs   \\
    GNAN \cite{GNAN} 
    & MICCAI'25 & Gaze   & 80.33~$\pm$~0.24 
    & 79.32~$\pm$~0.39  & 2.2 hrs   \\
    GradTrack \cite{GradTrack} 
    & MICCAI'25 & Gaze   & 80.25~$\pm$~0.40 
    & 81.01~$\pm$~0.66  & 2.2 hrs   \\
    \midrule
    \rowcolor{gray!25}
    \textbf{TrailNet (Ours)}
    & \textbf{Ours} & Gaze   & \textbf{81.25}~$\pm$~0.30 
    & \textbf{81.85}~$\pm$~0.39  & 2.2 hrs   \\
    
    \bottomrule
    \end{tabular}
    \caption{Comparison with different methods for five annotation types. Bold indicates the best results, and AT denotes the annotation time corresponding to each annotation type. TrailNet$_\text{F}$ represents the fully-supervised version of TrailNet.}
    \label{tab1}
\end{table*}

\subsection{Cycle Distillation Strategy (CDS)}
The CDS eliminates dependence on gaze input during inference by distilling trajectory-diagnostic priors from a teacher network into a student network, as illustrated in Fig.~\ref{fig_method2}. The teacher network $\mathcal{F}_\text{t}(\cdot, \cdot)$ consists of the full TSE encoder and MUD decoder. In contrast, the student network $\mathcal{F}_\text{s}(\cdot)$ only includes the vision encoder and MUD decoder and takes solely medical images as input.
Given input image $\mathbf{I}$, we apply strong augmentation $\mathcal{A}(\cdot)$ to obtain $\hat{\mathbf{I}} = \mathcal{A}(\mathbf{I})$, including random brightness and contrast adjustments. The original and augmented images are fed into the teacher and student networks respectively, yielding four multi-scale feature sets: teacher original features $\boldsymbol{\Phi}_\text{t}$, teacher augmented features $\hat{\boldsymbol{\Phi}}_\text{t}$, student original features $\boldsymbol{\Phi}_\text{s}$, and student augmented features $\hat{\boldsymbol{\Phi}}_\text{s}$. We construct cycle distillation constraints as:
\begin{equation}
\mathcal{L}_{\text{CDS}} = \frac{1}{4}(\mathcal{L}_{\text{c}}(\boldsymbol{\Phi}_\text{t}, \boldsymbol{\Phi}_\text{s}) + 
\mathcal{L}_{\text{c}} (\hat{\boldsymbol{\Phi}}_\text{t}, \hat{\boldsymbol{\Phi}}_\text{s}) + \mathcal{L}_{\text{c}}(\boldsymbol{\Phi}_\text{t}, \hat{\boldsymbol{\Phi}}_\text{t}) + \mathcal{L}_{\text{c}}(\boldsymbol{\Phi}_\text{s}, \hat{\boldsymbol{\Phi}}_\text{s})),
\end{equation}
where the cosine similarity loss between two arbitrary feature sets $\boldsymbol{\Phi}_a$ and $\boldsymbol{\Phi}_b$ is defined as:
\begin{equation}
\mathcal{L}_{\text{c}}(\boldsymbol{\Phi}_a, \boldsymbol{\Phi}_b) = \frac{1}{M} \sum_{m=1}^{M} \left(1 - \frac{\mathbf{F}_a^m \cdot \mathbf{F}_b^m}{\|\mathbf{F}_a^m\| \cdot \|\mathbf{F}_b^m\|} \right),
\end{equation}
with $M$ is the total number of feature levels, and $\mathbf{F}_a^m, \mathbf{F}_b^m$ denoting the $m$-th level feature maps from $\boldsymbol{\Phi}_a$ and $\boldsymbol{\Phi}_b$ respectively.

The overall training objective is:
\begin{equation}
\mathcal{L} = \mathcal{L}_{\text{seg}}(\mathbf{P}, \mathbf{L}) + \lambda_1~\mathcal{L}_{\text{MUD}} + \lambda_2~\mathcal{L}_{\text{CDS}},
\label{eq_loss}
\end{equation}
where $\mathcal{L}_{\text{seg}}(\mathbf{P}, \mathbf{L})$ denotes the supervised loss between full-resolution segmentation predictions and gaze pseudo-labels.

\section{Experiments}

\begin{figure}[t]
\centering
\includegraphics[width=0.99\columnwidth]{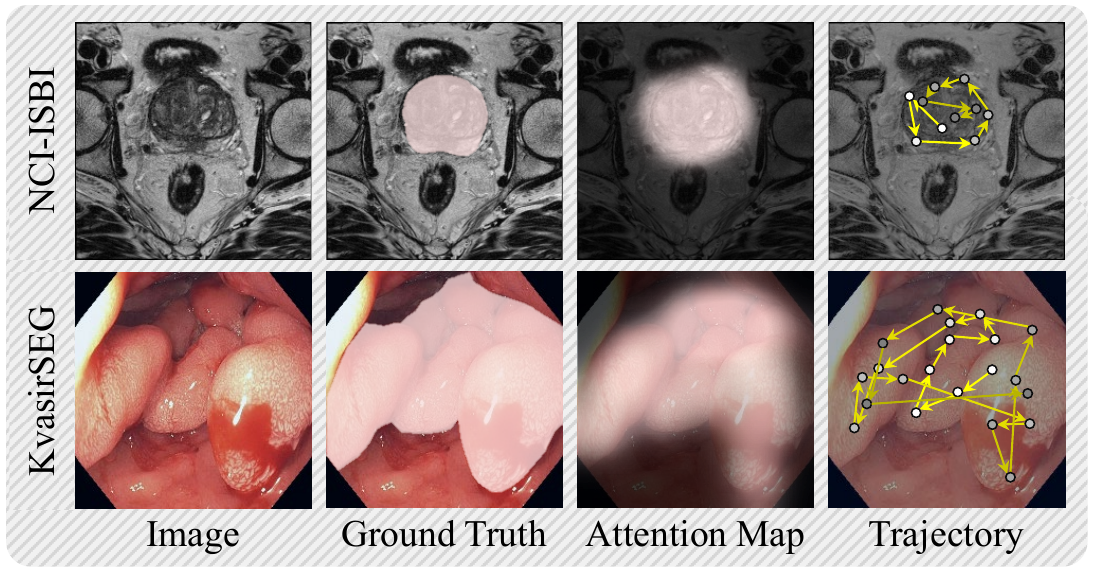}
\caption{Sample visualizations from the Kvasir-SEG and NCI-ISBI datasets. The fourth column shows trajectories, with dots denoting fixations and arrows indicating directions.}
\label{fig_dataset}
\end{figure}

Experiments are conducted on the public GazeMedSeg benchmark
\cite{zhong2024weakly}, which includes Kvasir-SEG \cite{Kvasir} and NCI-ISBI \cite{NCI}, as shown in Fig.~\ref{fig_dataset}. Kvasir-SEG \cite{Kvasir} is for gastrointestinal polyp segmentation and contains 900 training images and 100 test images. NCI-ISBI \cite{NCI} targets prostate segmentation from T2-weighted MRI and includes 789 training images and 117 test images. 
Details of gaze processing for pseudo-label and trajectory are provided in \textbf{Sec.~2} of the appendix.
Experiments were implemented in PyTorch on an NVIDIA 4090 GPU using the Adam optimizer for 100 epochs, with an initial learning rate of $1 \times 10^{-4}$ and a batch size of 4. The weighting coefficients $\lambda_1$ and $\lambda_2$ are both set to $1$. 
The foreground threshold $\tau_f$ and background threshold $\tau_b$ are set to $0.6$ and $0.3$, respectively \cite{GNAN}. The vision encoder employs Vision GNN \cite{ViG} pre-trained on ImageNet \cite{ImageNet}.
Studies of the threshold settings $\tau_f$ and $\tau_b$ and the backbone are provided in \textbf{Secs.~3} and \textbf{Secs.~4} of the appendix.
Segmentation performance is measured by the Dice coefficient, reported as the mean $\pm$ standard deviation over three independent runs.

\begin{figure*}[t]
\centering
\includegraphics[width=0.99\textwidth]{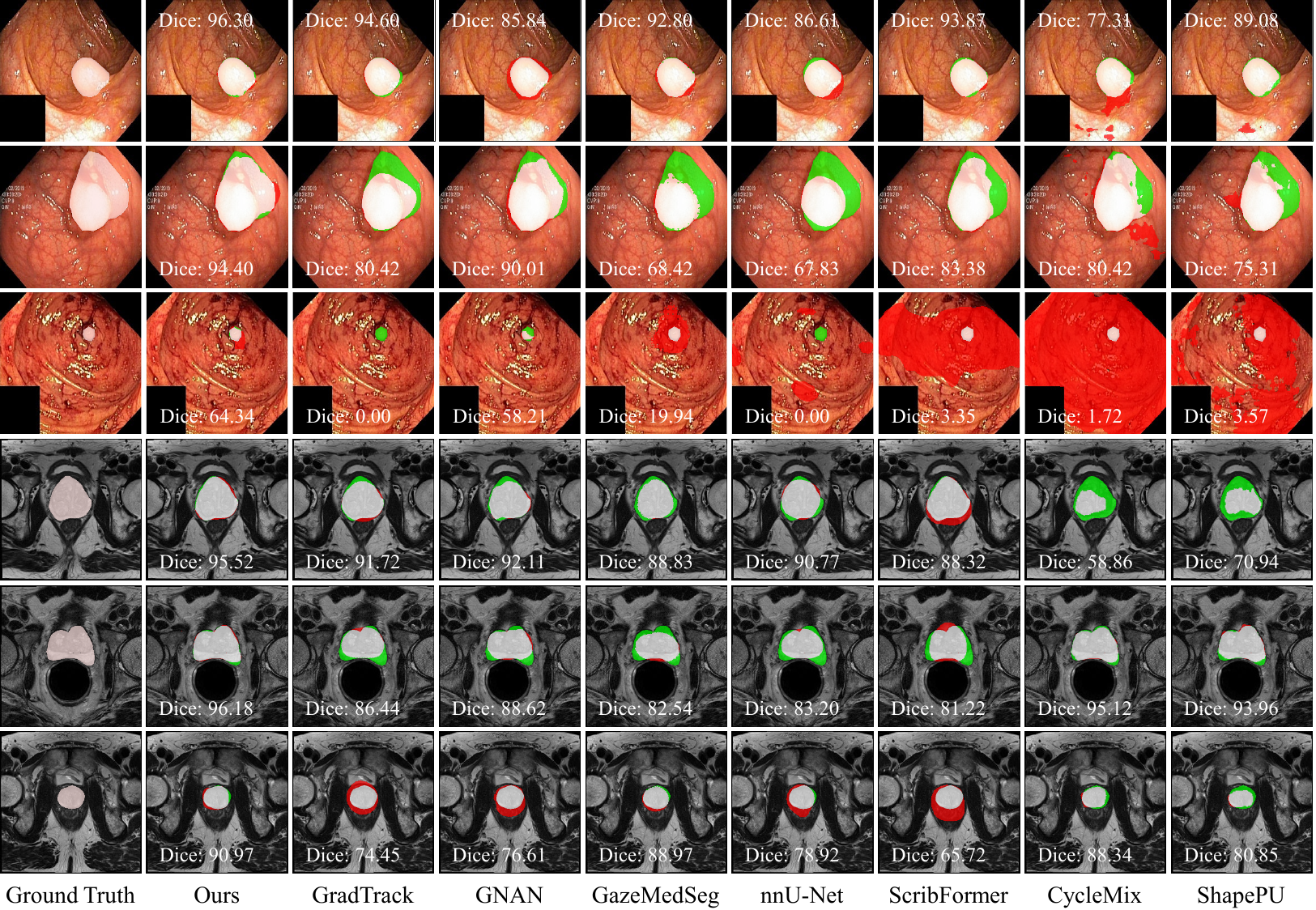}
\caption{Qualitative comparison of TrailNet with other state-of-the-art methods. Over-segmented areas are marked in red, and under-segmented areas are marked in green.}
\label{fig_result}
\end{figure*}

\subsection{Comparison with State-of-the-Art Methods}
\subsubsection{Quantitative Results.}
Table~\ref{tab1} reports the quantitative comparison across five supervision paradigms: 
full supervision, bounding-box, point, scribble, and gaze supervision.
TrailNet outperforms all other weakly supervised methods on both datasets, as confirmed by the Wilcoxon signed-rank test ($p<0.05$), achieving Dice scores of 81.25\% and 81.85\%, respectively.
Compared with the recent state-of-the-art (SOTA) method GradTrack, TrailNet improves Dice by 1.00\% (81.25\% vs. 80.25\%) on NCI-ISBI and 0.84\% (81.85\% vs. 81.01\%) on Kvasir-SEG. It also surpasses GNAN by 0.92\% and 2.53\% on the two datasets, respectively.
This performance gain can be attributed to the effective utilization of temporal trajectory information: our TSE extracts trajectory temporal context and MUD models uncertainty to mitigate pseudo-label noise.
Compared with ScribFormer, the SOTA scribble-supervised method, TrailNet achieves Dice improvements of 6.94\% and 6.16\% across the two datasets, demonstrating the stronger supervisory value of diagnostic trajectories.

TrailNet$_F$ denotes the fully supervised variant, where pseudo-labels $\mathbf{L}$ are replaced with dense ground truth. It outperforms nnU-Net by 1.24\% and 1.31\% on the two datasets, verifying that the TSE and MUD modules enhance feature representation for segmentation.
Under full supervision, Kvasir-SEG yields much higher Dice scores than NCI-ISBI due to differences in image properties and task difficulty: low-contrast MRI with ambiguous boundaries complicates prostate segmentation, whereas endoscopic polyps exhibit prominent contrast against surrounding tissues.

\subsubsection{Qualitative Results.}
Qualitative results are shown in Fig.~\ref{fig_result}, with over- and under-segmented regions highlighted in red and green, respectively. Compared with baselines, TrailNet produces more accurate boundaries by integrating spatial semantics with trajectory context through TSE and mitigating pseudo-label noise through MUD.

\begin{table}[t]
    \small
    \centering
    
    \begin{tabular}
    {m{1.3cm}<{\centering}  m{0.8cm}<{\centering} m{0.8cm}<{\centering} 
    | m{1.7cm}<{\centering} 
    m{1.7cm}<{\centering}}
    
    \toprule
    
    $\mathcal{L}_{\text{seg}}(\mathbf{P}, \mathbf{L})$
    & $\mathcal{L}_{\text{MUD}}$
    & $\mathcal{L}_{\text{CDS}}$
    & \textbf{NCI-ISBI} & \textbf{Kvasir-SEG}
    \\
    
    \midrule
    
    \checkmark & & & 
    78.96~$\pm$~0.44 & 79.81~$\pm$~0.48 \\
    \checkmark &\checkmark & &  
    80.54~$\pm$~0.39  & 81.28~$\pm$~0.36  \\
    \checkmark & &\checkmark &   
    80.69~$\pm$~0.42 & 81.33~$\pm$~0.40\\
    \midrule
    \rowcolor{gray!25}
    \checkmark &\checkmark &\checkmark &  
    \textbf{81.25}~$\pm$~0.30 & \textbf{81.85}~$\pm$~0.39\\
    
    \bottomrule
    \end{tabular}
    \caption{Ablation study on core loss functions. Bold indicates the best Dice scores.}
    
    \label{tab2}
\end{table}

\subsection{Ablation Study}
\subsubsection{Ablation of Loss Functions.}
Table~\ref{tab2} summarizes core module ablation results. The baseline is trained with gaze-supervised $\mathcal{L}_{\text{seg}}(\mathbf{P}, \mathbf{L})$ and basic teacher-student distillation loss $\mathcal{L}_{\text{c}}(\boldsymbol{\Phi}_t, \boldsymbol{\Phi}_s)$. 
Adding the MUD module yields consistent performance gains, indicating that explicit uncertainty modeling mitigates noisy pseudo-label interference. Further introducing the CDS brings additional improvements, validating the efficacy of cycle distillation constraints for feature distillation.
Table~\ref{tab3} reports ablation on loss terms within the MUD and CDS modules. For MUD, removing either the category mutual-exclusivity loss
$\mathcal{L}_u$ or the multi-scale supervision term $\mathcal{L}_m$ degrades performance, verifying the positive role of mutual-exclusivity regularization and multi-scale supervision in weakly supervised learning.
For CDS, removing the augmentation consistency constraints $\mathcal{L}_{\text{c}}(\boldsymbol{\Phi}_t, \hat{\boldsymbol{\Phi}}_t) + \mathcal{L}_{\text{c}}(\boldsymbol{\Phi}_s, \hat{\boldsymbol{\Phi}}_s)$ reduces segmentation accuracy, confirming the effectiveness of cross-view consistency regularization for robust prior distillation. Furthermore, CDS enables TrailNet (Student) to match the teacher's performance, verifying its effective transfer of trajectory knowledge.

\begin{table}[t]
    \small
    \centering
    
    \begin{tabular}
    {m{3.7cm}
    | m{1.7cm}<{\centering} 
    m{1.7cm}<{\centering}}
    
    \toprule
    
    \textbf{Method}
    & \textbf{NCI-ISBI} & \textbf{Kvasir-SEG}
    \\
    
    \midrule
    
    w/o $\mathcal{L}_{u}$  &
    80.89~$\pm$~0.33 & 81.41~$\pm$~0.48 \\
    w/o $\mathcal{L}_{m}$ &  
    80.94~$\pm$~0.42  & 81.47~$\pm$~0.44  \\
    w/o 
    $\mathcal{L}_{\text{c}}(\boldsymbol{\Phi}_\text{t}, \hat{\boldsymbol{\Phi}}_\text{t}), \mathcal{L}_{\text{c}}(\boldsymbol{\Phi}_\text{s}, \hat{\boldsymbol{\Phi}}_\text{s})$ 
    & 80.80~$\pm$~0.41 & 81.55~$\pm$~0.43\\
    \midrule

    TrailNet (Teacher)
    & 81.46~$\pm$~0.28 & 82.21~$\pm$~0.35\\

    \rowcolor{gray!25}
    TrailNet (Student) 
    & 81.25~$\pm$~0.30 & 81.85~$\pm$~0.39\\
    
    \bottomrule
    \end{tabular}
    \caption{Ablation study on loss terms within the MUD and CDS modules.}
    \label{tab3}
\end{table}

\begin{table}[t]
    \small
    \centering
    
    \begin{tabular}
    {m{3.7cm}
    | m{1.7cm}<{\centering} 
    m{1.7cm}<{\centering}}
    
    \toprule
    
    \textbf{Type of Trajectory}
    & \textbf{NCI-ISBI} & \textbf{Kvasir-SEG}
    \\
    
    \midrule
    Random (Gaze-level)  & 79.89~$\pm$~0.46 & 79.96~$\pm$~0.42 \\
    Random (Trajectory-level) & 80.35~$\pm$~0.36 & 80.79~$\pm$~0.45 \\
    Reversed Trajectory & 80.74~$\pm$~0.41  & 81.45~$\pm$~0.37  \\
    \midrule
    \rowcolor{gray!25}
    \textbf{Trajectory} 
    & \textbf{81.25}~$\pm$~0.30 & \textbf{81.85}~$\pm$~0.39\\
    
    \bottomrule
    \end{tabular}
    \caption{Ablation study on trajectory types. Bold denotes the best Dice scores.}
    \label{tab4}
\end{table}

\begin{table}[!t]
    \small
    \centering
    
    \begin{tabular}
    {m{3.7cm} | m{1.7cm}<{\centering} 
    m{1.7cm}<{\centering}}
    
    \toprule
    
    \textbf{Variant of the TSE}
    & \textbf{NCI-ISBI} & \textbf{Kvasir-SEG}
    \\
    
    \midrule
    DE (w/o $\textbf{f}_\text{sem}$) 
    & 80.72~$\pm$~0.26 & 81.38~$\pm$~0.38 \\
    DE (w/o $\textbf{f}_\text{phy}$) 
    & 80.83~$\pm$~0.39  & 81.43~$\pm$~0.40  \\
    TP (FFN) 
    & 80.59~$\pm$~0.44  & 81.15~$\pm$~0.46  \\
    TP (Transformer) 
    & 80.77~$\pm$~0.35  & 81.41~$\pm$~0.42  \\
    CI (Cross-Attention) 
    & 80.75~$\pm$~0.27  & 81.47~$\pm$~0.41  \\
    \midrule
    \rowcolor{gray!25}
    \textbf{TrailNet} 
    & \textbf{81.25}~$\pm$~0.30 & \textbf{81.85}~$\pm$~0.39\\
    
    \bottomrule
    \end{tabular}
    \caption{Ablation study on block architectures within the TSE module. Bold indicates the best Dice scores.}
    \label{tab5}
\end{table}

\subsubsection{Effectiveness of the TSE.}
\begin{figure}[t]
\centering
\includegraphics[width=0.99\columnwidth]{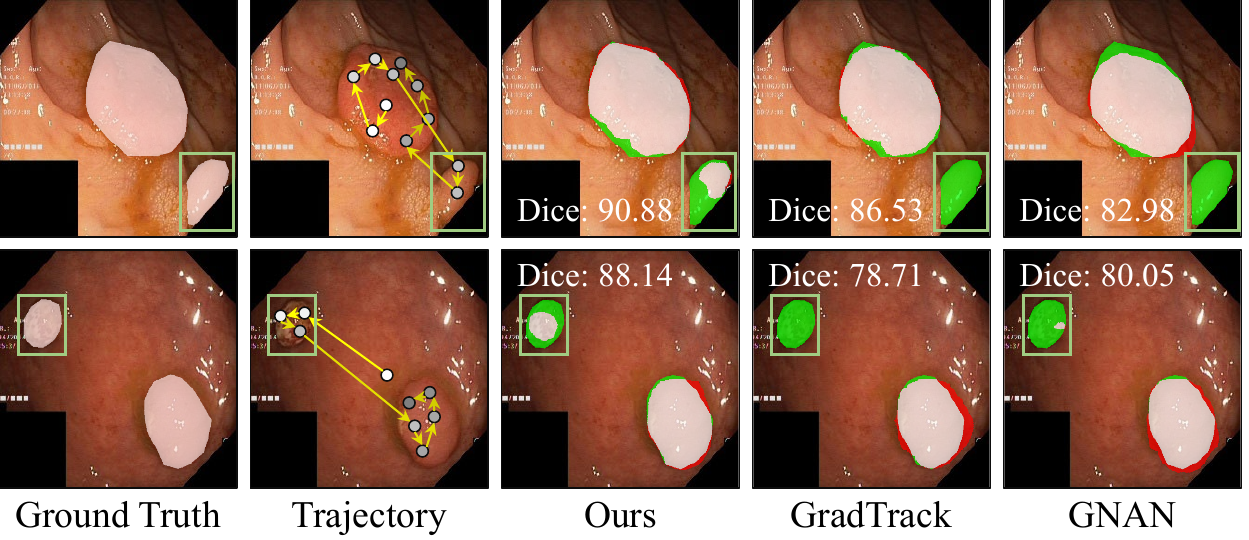}
\caption{Visualization of the contribution of trajectories in segmentation. Methods without explicit temporal trajectory modeling fail to detect tiny polyps (marked by green boxes), while the proposed method accurately identifies these lesions under trajectory guidance.}
\label{fig_ablation}
\end{figure}

Table~\ref{tab4} evaluates the effectiveness of diagnostic trajectory priors. Randomizing fixation locations, shuffling fixation order, or reversing trajectories consistently reduces Dice scores, verifying that both spatial locations and temporal order of gaze trajectories are vital for accurate segmentation.
As shown in Fig.~\ref{fig_ablation}, methods without explicit temporal trajectory modeling may miss tiny polyps, whereas TrailNet accurately identifies them using clinical viewing patterns.

Table~\ref{tab5} further reports ablation studies on internal blocks of the TSE module. Removing either the image semantic features $\mathbf{f}_{\text{sem}}$ or the physical features $\mathbf{f}_{\text{phy}}$ from the DE module reduces performance, indicating that complementary semantic and physical attributes jointly provide more comprehensive trajectory representations. Replacing the TP block with FFN or Transformer layers degrades results, validating the advantage of selective state space models for long-range temporal modeling. Replacing the CI module with vanilla cross-attention also lowers performance, demonstrating the superiority of the context-semantic interaction block.

\subsubsection{Analysis of the Loss Weights.}
\begin{figure}[!t]
    \centering
    \begin{subfigure}[b]{0.495\columnwidth}
        \centering
        \includegraphics[width=\linewidth]{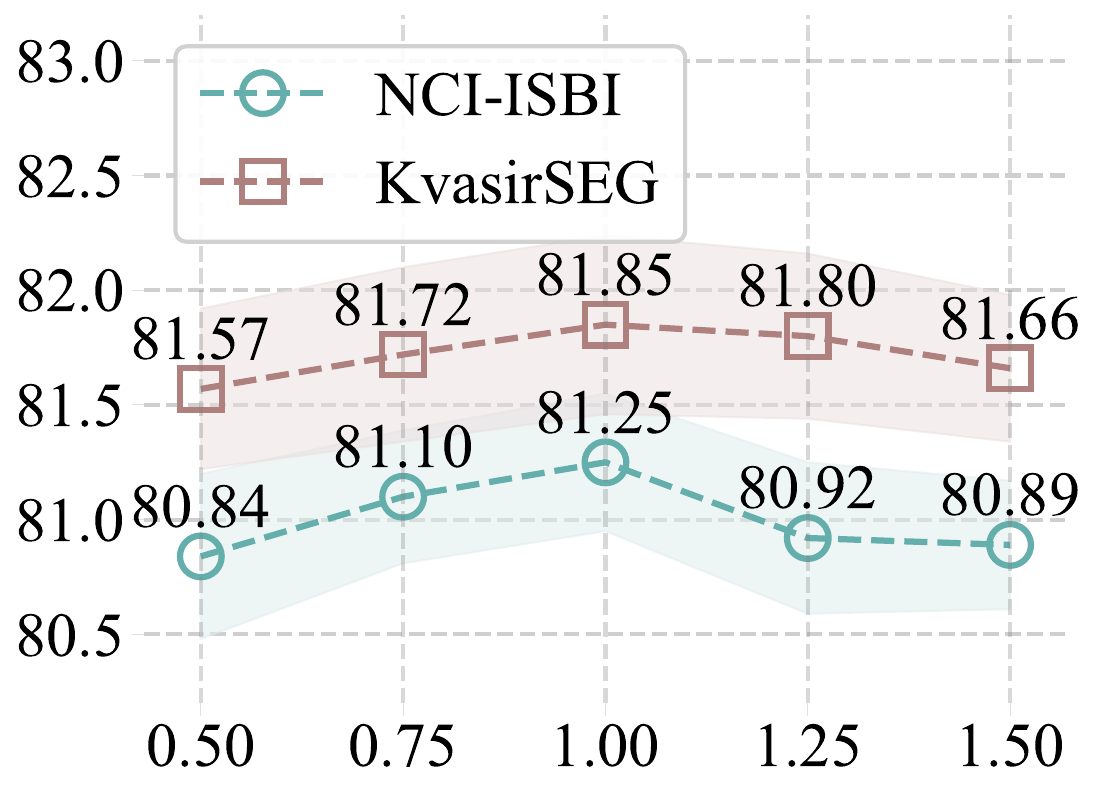}
        \caption{$\lambda_1$ ($\lambda_2=1.0$)}
        \label{fig:left}
    \end{subfigure}
    \hfill
    \begin{subfigure}[b]{0.495\columnwidth}
        \centering
        \includegraphics[width=\linewidth]{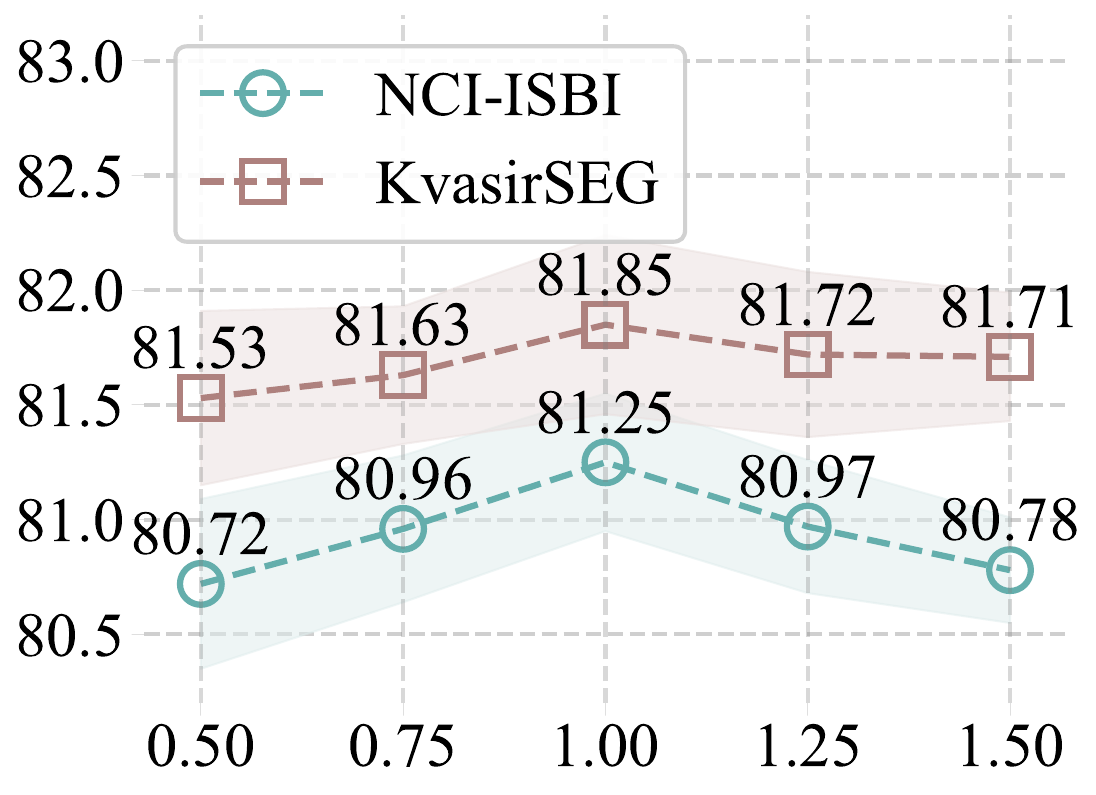}
        \caption{$\lambda_2$ ($\lambda_1=1.0$)}
        \label{fig:right}
    \end{subfigure}
    \caption{Parameter study on $\lambda_1$ and $\lambda_2$. The vertical axis indicates the Dice score.}
    \label{fig_lambda}
\end{figure}

The loss function of our TrailNet involves two hyperparameters, as formulated in Eq.~(\ref{eq_loss}). $\lambda_1$ controls the weight of $\mathcal{L}_{\text{MUD}}$, while $\lambda_2$ controls that of $\mathcal{L}_{\text{CDS}}$.
To analyze parameter sensitivity, each coefficient is varied over five values, \textit{i.e.}, $\{0.5, 0.75, 1.0, 1.25, 1.5\}$. 
As illustrated in Fig.~\ref{fig_lambda}, the model performance fluctuates slightly with different settings of $\lambda_1$ and $\lambda_2$ while remaining stable overall.

\section{Conclusion}
In this paper, we propose TrailNet that exploits the temporal context of trajectories to inject diagnostic-related visual patterns from clinical image reading into gaze-supervised medical image segmentation. TrailNet comprises three core modules. The TSE models temporal context from trajectories and establishes complementary interactions with image spatial semantics to strengthen target perception. The MUD drives the model to learn deterministic predictions via category mutual-exclusivity constraints, mitigating the adverse effects caused by gaze noise. Finally, the CDS transfers feature-level knowledge via teacher-student networks and cycle distillation constraints to enable gaze-free inference.  
Experimental results on two public datasets show that TrailNet outperforms SOTA weakly supervised segmentation methods, validating that trajectory-embedded temporal context boosts segmentation performance. Future work will extend TrailNet to a broader range of medical imaging modalities and generalize gaze supervision to diverse clinical scenarios.

\bibliography{aaai2027}


\end{document}